\documentclass{article} 
\usepackage{iclr2020_conference,times}

\iclrfinalcopy

\usepackage{amsmath,amsfonts,bm}

\def\eqref#1{equation~\ref{#1}}

\def\1{\bm{1}}

\def\rz{{\textnormal{z}}}

\def\rvm{{\mathbf{m}}}

\def\rvx{{\mathbf{x}}}

\def\rvz{{\mathbf{z}}}

\def\rmI{{\mathbf{I}}}

\def\rmM{{\mathbf{M}}}

\DeclareMathAlphabet{\mathsfit}{\encodingdefault}{\sfdefault}{m}{sl}
\SetMathAlphabet{\mathsfit}{bold}{\encodingdefault}{\sfdefault}{bx}{n}

\newcommand{\E}{\mathbb{E}}
\newcommand{\Ls}{\mathcal{L}}

\newcommand{\KL}{D_{\mathrm{KL}}}

\usepackage{hyperref}
\usepackage{url}
\usepackage{booktabs}
\usepackage{graphicx}
\usepackage[font=small,labelfont=bf]{caption}

\title{Generative Hierarchical Models for Parts, Objects, and Scenes}

\author{Fei Deng\\
Department of Computer Science\\
Rutgers University\\
\texttt{fei.deng@rutgers.edu}
\And
Zhuo Zhi\thanks{Work done while visiting Rutgers University}\\
Department of Electrical Engineering\\
Xi'an Jiaotong University\\
\texttt{zhizz001@stu.xjtu.edu.cn} \\
\AND
Sungjin Ahn \\
Department of Computer Science\\
Rutgers University \\
\texttt{sungjin.ahn@rutgers.edu}
}

\newcommand{\pres}{\text{pres}}
\newcommand{\pose}{\text{pose}}
\newcommand{\appr}{\text{appr}}
\newcommand{\addr}{\text{addr}}
\newcommand{\what}{\text{what}}
\newcommand{\ratio}{\text{ratio}}
\newcommand{\where}{\text{where}}

\begin{document}

\maketitle

\begin{abstract}

Compositional structures between parts and objects are inherent in natural scenes. Modeling such compositional hierarchies via unsupervised learning can bring various benefits such as interpretability and transferability, which are important in many downstream tasks. In this paper, we propose the first deep latent variable model, called RICH, for learning Representation of Interpretable Compositional Hierarchies.~At the core of RICH is a latent scene graph representation that organizes the entities of a scene into a tree structure according to their compositional relationships. During inference, taking top-down approach, RICH is able to use higher-level representation to guide lower-level decomposition. This avoids the difficult problem of routing between parts and objects that is faced by bottom-up approaches. In experiments on images containing multiple objects with different part compositions, we demonstrate that RICH is able to learn the latent compositional hierarchy and generate imaginary scenes.

\end{abstract}

\section{Introduction}

Compositional hierarchies prevail in natural scenes where primitive entities are recursively composed into more abstract entities. Modeling such compositional generative process allows discovery of modular primitives that can be reused across a variety of scenes. Hence, it would bring interpretability and transferability, in which current deep learning models are not quite successful. Due to expensive labeling, such compositional relationships should ideally be learned in an unsupervised manner. Unsupervised approaches can also provide more flexibility and generalization ability since the model is allowed to choose the most appropriate compositional hierarchy for a given scenario.

Despite its importance, there has not been much work on unsupervised generative modeling of the compositional hierarchy. Earlier work on hierarchical representation learning \citep{lee2009convolutional} obtains a feature hierarchy that captures concepts at different levels of abstraction, with no explicit modeling of composition. Recent researches on deep latent variable models \citep{maaloe2019biva, zhao2017learning, sonderby2016ladder, bachman2016architecture} mainly focus on architectural designs and training methods that harness the full expressive power of hierarchical generative models. Although they have shown impressive generation quality and disentanglement of learned representation, the compositional hierarchy is still not captured in a modular and interpretable way. To obtain interpretable scene representation, recent work \citep{tieleman2014optimizing, eslami2016attend, crawford2019spatially, wu2017neural, yao20183d, romaszko2017vision, deng2019cerberus} has introduced domain-specific decoders that take object pose and appearance information as input and render the object in a similar way to graphics engines. This forces the encoder to invert the rendering process, producing interpretable object-wise pose and appearance representation.

In this paper, we extend the interpretable object-wise representation to the hierarchical setting. We propose a deep generative model, called RICH (Representation of Interpretable Compositional Hierarchies), that can use its hierarchy to represent the compositional relationships among interpretable symbolic entities like parts, objects, and scenes. To this end, taking inspiration from capsule networks \citep{sabour2017dynamic, e2018matrix} and the rendering process of computer graphics, we propose a \textit{probabilistic scene graph representation} that describes the compositional hierarchy as a latent tree. The nodes in the tree correspond to entities in the scene, while the edges indicate the compositional relationships among these entities. We associate an appearance latent with each node to summarize all lower-level composition, and a pose latent with each edge to specify the transformation from the current level to the upper level. To enforce interpretability, the probabilistic scene graph is then paired with a decoder that renders the scene graph by recursively applying the specified transformations. We also introduce learnable templates for the primitive entities. Once learned, RICH is able to generate all lower-level latents and render a partial scene given the latent at a specific level. 

To infer the scene graph is, however, challenging, since both the tree structure and the latent variables need to be simultaneously inferred. Capsule networks have provided a bottom-up solution to learning the tree structure, but it faces the difficult routing problem caused by the exponentially many possible compositions. Instead, RICH takes a top-down approach that avoids the routing problem. The intuition is that for a given scene, it is natural to first decompose it into high-level objects. If we devote our attention to one of the objects, we can then figure out its constituent parts. In cases where parts are close or have occlusion,  we expect the appearance latent of the higher-level object to guide lower-level decomposition, since it summarizes the typical composition for that object.

The contributions of this paper are as follows. We propose RICH, the first interpretable representation learning model for compositional hierarchies through probabilistic latent variable modeling. We then implement a three-level prototype of RICH and demonstrate its effectiveness in extensive experiments. RICH is able to learn the hierarchical scene graph representation from images containing multiple compositional objects. Further, it shows decent generation quality and generalization ability to unseen number of objects.

\section{Related work}

\textbf{Interpretable object-wise representation.} AIR \citep{eslami2016attend} is the first generative model that learns interpretable object-wise scene representation. It is able to assign a latent vector $(\rz^\pres, \rvz^\where, \rvz^\what)$ to each object in the scene, describing the presence, size, center position, and appearance of the object. SPAIR \citep{crawford2019spatially} improves the scalability of AIR to images containing a large number of objects. It divides the image into spatially distributed cells, and auto-regressively infers the latent vector for each cell. This crucially reduces the search space for individual cells, since they are each responsible only for explaining objects near themselves. RICH builds upon SPAIR to infer the structure of the probabilistic scene graph. To enable efficient hierarchical inference, we use mean-field approximation for the posterior, allowing inference of all cells to be done in parallel. 

\textbf{Hierarchical scene representation.} Modeling the part-whole relationship in scenes has attracted growing interest, and it has been utilized for improving image classification, parsing, and segmentation. Two representative models are hierarchical compositional models (HCMs) and capsule networks. In HCMs \citep{zhu2008unsupervised}, the hierarchical structure is represented as a graph, where leaf nodes interact with image segments, and upper-level nodes store the average position and orientation of lower-level nodes (with respect to the image coordinates). In capsule networks \citep{sabour2017dynamic, e2018matrix}, the part-whole relationship is used for achieving viewpoint invariance. The key insight is that the relative pose of parts with respect to objects is viewpoint invariant, and is thus suitable to be learned as network weights. However, neither of these two approaches uses generative modeling, and they have been applied only to scenes with one dominant object.

\section{The proposed model: RICH}

RICH (Representation of Interpretable Compositional Hierarchies) is a generative model that captures the recursive compositional structure inherent in natural scenes. It builds a tree-structured representation similar to scene graphs in computer graphics \citep{foley1996computer}. The nodes in the tree describe entities at various levels of abstraction in the scene, and the edges indicate the compositional relationships among these entities. Specifically, each leaf node represents a primitive entity that is not further decomposed. Each internal node represents an abstract entity that is composed from its children nodes. The composition is specified by the relative pose of each child node with respect to the parent node, and this pose information is stored on the corresponding edges.

\subsection{Generative process}

\begin{figure}[t]
    \centering \includegraphics[width=0.95\columnwidth]{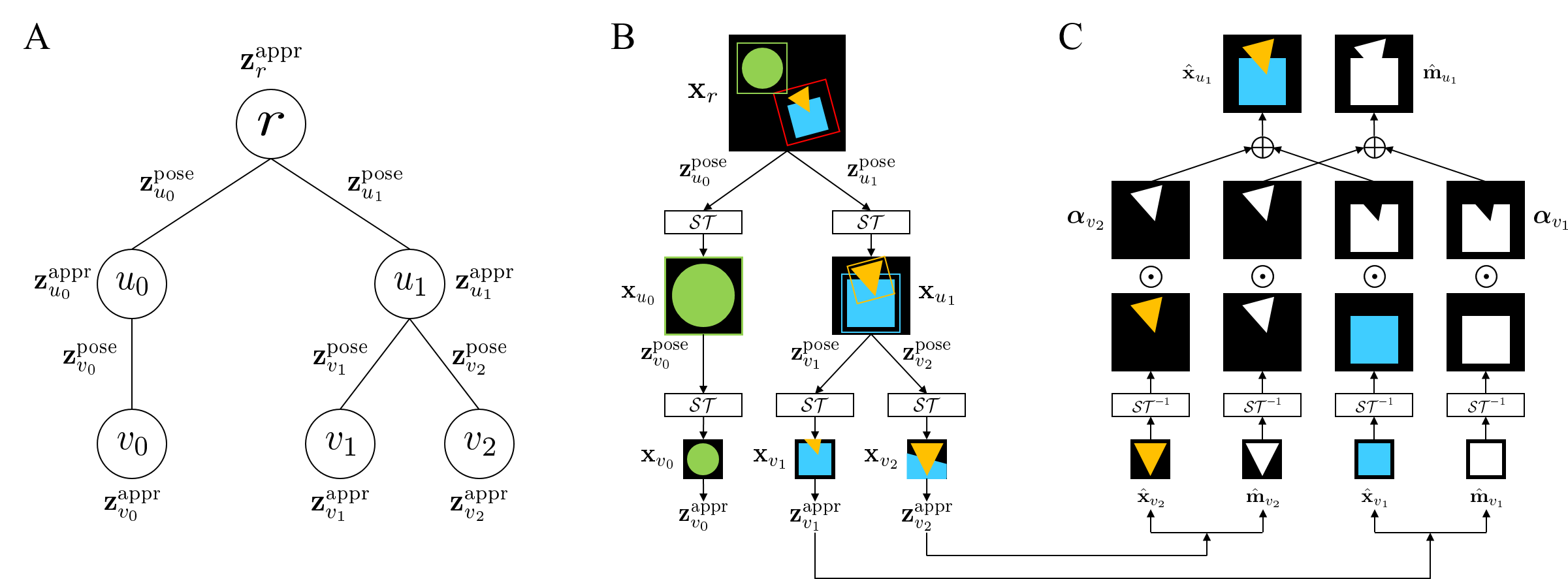}
    \caption{{\em (A)} Probabilistic scene graph representation. Each node represents an entity in the scene, and is associated with an appearance latent. Each edge is associated with a relative pose latent that specifies the coordinate transformation between the child node and the parent node. {\em (B)} Top-down inference process. Inference combines information from glimpse regions and higher-level appearance latents (not shown here). Bounding boxes indicate inferred pose latents. {\em (C)} Recursive decoding process (a single recursive step shown). The image patch $\hat{\rvx}_{u_1}$ and mask $\hat{\rvm}_{u_1}$ of an internal node $u_1$ are decoded from the image patches and masks of all its children nodes.}
    \label{fig:model}
\end{figure}

To make a generative model, we associate an appearance vector $\rvz_v^\appr$ with each node $v$, and a pose vector $\rvz_v^\pose$ with the edge between node $v$ and its parent $pa(v)$, as shown in Figure \ref{fig:model}A. The intuition is that $\rvz_{pa(v)}^\appr$ represents the entity at $pa(v)$ in its canonical pose, summarizing all lower-level composition in the subtree rooted at $pa(v)$. Conditioning on $\rvz_{pa(v)}^\appr$, we can then sample $\rvz_v^\appr$ and $\rvz_v^\pose$, the relative pose of $v$ with respect to $pa(v)$. Thus, the latent vectors in any subtree can be recursively generated. Let $V$ be the set of all nodes, $r \in V$ be the root node, and $L \subseteq V$ be the set of leaf nodes. The generative model for the entire scene $\rvx$ can be written as follows:
\begin{equation}
    p(\rvx) = \int p(\rvx \mid \rvz_{V \setminus \{r\}}^\pose, \rvz_L^\appr) \, p(\rvz_r^\appr) \prod_{v \in {V \setminus \{r\}}} p(\rvz_v^\pose \mid \rvz_{pa(v)}^\appr) \, p(\rvz_v^\appr \mid \rvz_{pa(v)}^\appr) \ d\rvz,
    \label{eqn:generative}
\end{equation}
where we assume conditional independence among all latents $\rvz_v^\pose$ and $\rvz_v^\appr$ that have the same $pa(v)$. This gives disentangled and interpretable scene representation.

We design the decoder $p(\rvx | \rvz_{V \setminus \{r\}}^\pose, \rvz_L^\appr)$ to closely follow the rendering process from a given scene graph. First, for each leaf node $v \in L$, we use a neural network to decode its appearance vector into a small image patch $\hat{\rvx}_v$ and a (close to) binary mask $\hat{\rvm}_v$ the same size as $\hat{\rvx}_v$. Here we assume that $\hat{\rvx}_v$ has already been masked by $\hat{\rvm}_v$, meaning $\hat{\rvx}_v(i,j)=0$ for all pixel locations $(i,j)$ where $\hat{\rvm}_v(i,j)=0$. We then recursively compose these primitive patches into the entire scene by applying affine transformations level by level. Specifically, let $u$ be an internal node, and $ch(u)$ be the set of its children. We compose the higher-level image patch $\hat{\rvx}_u$ and mask $\hat{\rvm}_u$ as follows:
\begin{equation}
    \hat{\rvx}_u = \sum_{v \in ch(u)} \bm{\alpha}_v \odot \mathcal{ST}^{-1}(\hat{\rvx}_v, \ \rvz_v^\pose), \quad \ 
    \hat{\rvm}_u = \sum_{v \in ch(u)} \bm{\alpha}_v \odot \mathcal{ST}^{-1}(\hat{\rvm}_v, \ \rvz_v^\pose),
    \label{eqn:decoder}
\end{equation}
where a spatial transformer $\mathcal{ST}$ \citep{jaderberg2015spatial} is used to properly place $\hat{\rvx}_v$ and $\hat{\rvm}_v$ into the coordinate frame of the parent node $u$, according to the scaling, rotation and translation parameters given by $\rvz_v^\pose$. In addition, $\rvz_v^\pose$ also provides relative depth information that helps deal with occlusion. Entities with smaller depth will appear in front of entities with larger depth. This is enforced by a transparency map $\bm{\alpha}_v$ that assigns pixel-wise weights to each transformed patch according to its relative depth. See Figure \ref{fig:model}C for an illustration. To ensure that unoccluded part of entities will remain visible, we normalize $\bm{\alpha}_v$ after applying the transformed mask $\tilde{\rvm}_v = \mathcal{ST}^{-1}(\hat{\rvm}_v, \ \rvz_v^\pose)$, namely for all pixel locations $(i, j)$,
\begin{equation}
    \bm{\alpha}_v(i, j) = 0 \ \ \text{if} \ \ \tilde{\rvm}_v(i,j) = 0, \quad
    \sum_{v \in ch(u)} \bm{\alpha}_v(i, j) = 1 \ \ \text{if} \sum_{v \in ch(u)} \tilde{\rvm}_v(i,j) > 0.
    \label{eqn:transparency}
\end{equation}
The final decoder output $p(\rvx | \rvz_{V \setminus \{r\}}^\pose, \rvz_L^\appr) = \mathcal{N}(\hat{\rvx}_r, \sigma^2\rmI)$, where $\sigma$ is a hyperparameter.

\subsection{Inference and learning}

Since computing $p(\rvx)$ in Equation~\ref{eqn:generative} is intractable, we train RICH with variational inference. The approximate posterior is designed to factorize $\rvz$ in the top-down fashion similar to the generative process, such that a higher-level appearance representation guides lower-level decomposition:
\begin{equation}
    p(\rvz \mid \rvx) \approx q(\rvz \mid \rvx) = q(\rvz_r^\appr \mid \rvx) \prod_{v \in {V \setminus \{r\}}} q(\rvz_v^\pose \mid \rvz_{pa(v)}^\appr, \rvx_{pa(v)}) \, q(\rvz_v^\appr \mid \rvz_{pa(v)}^\appr, \rvx_v).
    \label{eqn:posterior}
\end{equation}
Here $\rvx_v$ is the region of input image $\rvx$ that corresponds to the entity that node $v$ represents. This region is specified by all the predicted pose vectors along the path from root $r$ to node $v$. More precisely, we define $\rvx_r = \rvx$, and recursively extract $\rvx_v = \mathcal{ST}(\rvx_{pa(v)}, \rvz_v^\pose)$, as shown in Figure \ref{fig:model}B. Notice that the relative pose $\rvz_v^\pose$ of $v$ with respect to $pa(v)$ needs to be inferred from $\rvx_{pa(v)}$. After applying spatial transformer $\mathcal{ST}$, we assume the captured $\rvx_v$ is in its canonical pose. This top-down inference process avoids the challenging routing problem in capsule networks \citep{sabour2017dynamic, e2018matrix}, leading to more efficient inference. In cases where entities are close or have occlusion, the higher-level appearance vector $\rvz_{pa(v)}^\appr$ could provide some guidance on separating these entities.

In general, all latents are assumed to be continuous, with both prior and posterior being Gaussian distributions. However, it may bring additional flexibility and interpretability to introduce some discrete latents, as we explain in Section \ref{sec:detail}. For continuous latents, we compute posterior via precision-weighted combination similar to Ladder-VAE \citep{sonderby2016ladder}, and use reparameterization trick \citep{kingma2013auto} to sample from the posterior. For discrete latents, we use Gumbel-Softmax trick \citep{jang2016categorical, maddison2016concrete}. Thus, the entire model can be trained end-to-end via backpropagation to maximize the following evidence lower bound (ELBO):
\begin{align}
    \Ls &= \E_{q(\rvz \mid \rvx)} [\log p(\rvx \mid \rvz_{V \setminus \{r\}}^\pose, \rvz_L^\appr)]
           - \KL[q(\rvz_r^\appr \mid \rvx) \parallel p(\rvz_r^\appr)] \label{eqn:elbo} \\ \notag
        &\phantom{{}={}} - \sum\nolimits_{v \in {V \setminus \{r\}}}
           \E_{q(\rvz \mid \rvx)} [\KL[q(\rvz_v^\pose \mid \rvz_{pa(v)}^\appr, \rvx_{pa(v)}) \parallel p(\rvz_v^\pose \mid \rvz_{pa(v)}^\appr)]] \\ \notag
        &\phantom{{}={}} - \sum\nolimits_{v \in {V \setminus \{r\}}}
           \E_{q(\rvz \mid \rvx)} [\KL[q(\rvz_v^\appr \mid \rvz_{pa(v)}^\appr, \rvx_v) \parallel p(\rvz_v^\appr \mid \rvz_{pa(v)}^\appr)]].
\end{align}

\subsection{Implementation details} \label{sec:detail}

\textbf{Structural inference.} In our description above, we have assumed that the tree structure is already known. We now relax this assumption and introduce structural inference. First, we set a maximum out-degree for each node so that the number of all possible structures is bounded. For simplicity, in our implementation nodes within one level share the same maximum out-degree. To determine the structure, it then suffices to specify the presence of each possible edge. Hence, for an arbitrary edge between node $v$ and its parent, we introduce a Bernoulli variable $\rz_v^\pres$ to indicate its presence. If $\rz_v^\pres = 0$, meaning the edge is not present, then $\rvz_v^\pose$ along with all latents in the subtree rooted at $v$ are excluded from the representation. To encourage sparse structures, we initialize the prior $p(\rz_v^\pres \mid \rvz_{pa(v)}^\appr)$ and the posterior $q(\rz_v^\pres \mid \rvz_{pa(v)}^\appr, \rvx_{pa(v)})$ to have small Bernoulli parameters.

\textbf{Node grounding.} Due to the symmetric tree structure, there are numerous equivalent entity-to-node assignments for a given scene, each yielding a different permutation of the pose vectors. This can cause difficulties in the learning process. In particular, the model has to learn a consistent assignment strategy such that the pose vector at each edge can be well captured by a unimodal Gaussian distribution. To alleviate this problem, we impose some inductive bias on the assignment strategy. Inspired by SPAIR \citep{crawford2019spatially}, for each internal node $u$, we divide $\rvx_u$ into a grid of $N_u$ cells, where $N_u$ is the maximum out-degree of $u$. Each child of $u$ is assigned to one of these cells, and is responsible for explaining only the entity whose center position is within that cell. Assuming that $\rvx_u$ captures the entity $u$ in its canonical pose, this assignment strategy ensures that each child of node $u$ is almost always associated with the same component of entity $u$. To deal with occlusion, we let neighboring cells have slight overlap.

\textbf{Primitive templates.} It is often reasonable to assume that the vast number of complex entities can be composed from only a modest number of primitive entities. Identifying such primitive entities through discrete latent variables would bring additional interpretability. Hence, we introduce an external memory $\rmM$ with variational addressing \citep{bornschein2017variational} to learn a set of primitive templates. For each leaf node $v \in L$, we decompose its appearance vector into $\rvz_v^\appr = (\rvz_v^\addr, \rvz_v^\what)$. Here, $\rvz_v^\addr$ is a one-hot vector that points one of the templates in $\rmM$, and $\rvz_v^\what$ is a continuous vector that explains the remaining variability. We assume that $\rvz_{pa(v)}^\appr$ captures the identity but not the exact appearance of entity $v$, since $\rvz_{pa(v)}^\appr$ is intended for summarization. Therefore, we factorize the prior and the posterior as follows:
\begin{align}
    p(\rvz_v^\appr \mid \rvz_{pa(v)}^\appr) &= p(\rvz_v^\addr \mid \rvz_{pa(v)}^\appr, \rmM) \, p(\rvz_v^\what \mid \rmM[\rvz_v^\addr]) , \\
    q(\rvz_v^\appr \mid \rvz_{pa(v)}^\appr, \rvx_v) &= q(\rvz_v^\addr \mid \rvz_{pa(v)}^\appr, \rmM, \rvx_v) \, q(\rvz_v^\what \mid \rmM[\rvz_v^\addr], \rvx_v) ,
\end{align}
where $\rmM$ is considered as model parameter, and $\rmM[\rvz_v^\addr]$ is the deterministically retrieved memory content.
We implement $\rmM$ to be a stack of low-dimensional embeddings of templates. To decode from $\rvz_v^\appr$, we first retrieve the embedding indexed by $\rvz_v^\addr$, and decode it into a single-channel image patch. This serves as both the template and the mask, namely $\hat{\rvm}_v = g(\rmM[\rvz_v^\addr])$. We then apply multiplicative modification controlled by $\rvz_v^\what$ and obtain $\hat{\rvx}_v = \hat{\rvm}_v \odot h(\rvz_v^\what)$. Here, $\hat{\rvx}_v$ has the same number of channels as the input $\rvx$, and both $g(\cdot)$ and $h(\cdot)$ are implemented as spatial broadcast decoders \citep{watters2019spatial}.

\textbf{Canonical size.} To avoid undesired effects in cascaded affine transformations, we constrain the spatial transformer to always preserve aspect ratio. Thus, for a square input image, each entity is assumed to occupy a square image region. Considering the various possible compositions, this region may not capture the entity's canonical size well. Hence, we introduce a latent variable $\rz_u^\ratio$ to represent the aspect ratio of entity $u$. This could give a tighter bounding box inside the square region. We consider $\rz_u^\ratio$ as part of the appearance vector, namely $\rvz_u^\appr = (\rvz_u^\what, \rz_u^\ratio)$, and factorize the prior and the posterior as follows:
\begin{align}
    p(\rvz_u^\appr \mid \rvz_{pa(u)}^\appr) &= p(\rvz_u^\what \mid \rvz_{pa(u)}^\appr) \, p(\rz_u^\ratio \mid \rvz_u^\what) , \\
    q(\rvz_u^\appr \mid \rvz_{pa(u)}^\appr, \rvx_u) &= q(\rvz_u^\what \mid \rvz_{pa(u)}^\appr, \rvx_u) \, q(\rz_u^\ratio \mid \rvz_u^\what, \rvx_u) .
\end{align}
For simplicity, we introduce $\rz_u^\ratio$ only for intermediate nodes $u \in {V \setminus (L \cup \{r\})}$. To properly learn $\rz_u^\ratio$, we feed it as an additional argument to the spatial transformer. This has two effects. First, the region outside the bounding box given by $\rz_u^\ratio$ is masked out during transformation in both inference and reconstruction. This forces the bounding box to capture the entity in its entirety. Second, during training, we inject zero-mean Gaussian noise to the reconstruction inside the bounding box region. The noise level is annealed as training proceeds. This encourages tight bounding boxes without affecting generation quality.

\section{Experiments}

\subsection{Datasets and a three-level prototype}

We have implemented a prototype of RICH with part-, object-, and scene-level representation. We will refer to the latents as $\rvz_P$, $\rvz_O$, and $\rvz_S$ respectively. The maximum out-degree is set to be 4 for each internal node. For evaluation, we have made two datasets of 2D and 3D scenes. Both datasets contain 128$\times$128 color images, split into 64000 for training, 12800 for validation, and 12800 for testing. They present challenges of (i) multi-pose, variable number of objects and parts, (ii) multiple occurrences of the same type of objects and parts within one scene, and (iii) severe occlusion in 3D scenes. In making each dataset, we first choose a set of primitive shapes to be the parts, and then construct the objects and scenes by recursively composing these parts. Specifically, we have chosen three shapes as parts, and defined ten types of objects in terms of the identity of the constituent parts and their relative position, scale, and orientation. Among these ten types, three contain a single part, another three contain two parts, and the remaining four contain three parts. To construct a scene, we first randomly sample the number of objects (between one and four) and their types, and then instantiate these objects. This means for each object, we choose a random color for each of its parts, apply random scaling (within 10\% of object size), and draw it at a random position in the scene. In 2D case, the instantiation process also includes random perturbation of parts and random rotation of the object as a whole. We ensure that different objects have minimal overlap. In 3D case, we use MuJoCo \citep{todorov2012mujoco} to place the objects on a plane, and then take observations from ten different viewpoints, some of which can lead to severe occlusion.

\begin{figure}[t]
    \centering \includegraphics[width=0.95\columnwidth]{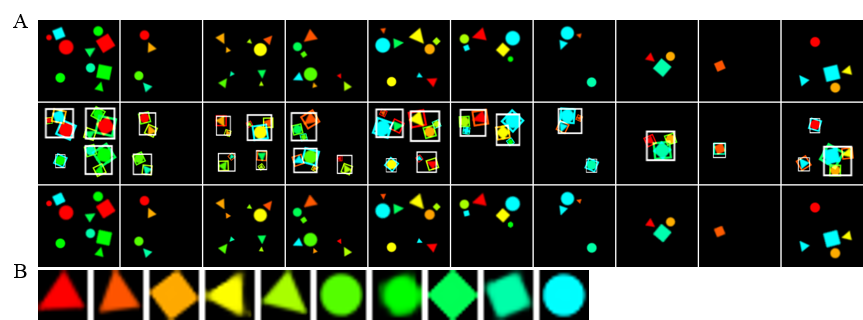}
    \caption{Qualitative results on 2D dataset. {\em (A) (Top)} Input image. {\em (Middle)} Input image superimposed with predicted bounding boxes, drawn according to $\rvz_O^\pres$, $\rvz_O^\pose$, $\rvz_O^\ratio$, $\rvz_P^\pres$, $\rvz_P^\pose$ and $\rvz_P^\addr$. {\em (Bottom)} Reconstruction. {\em (B)} Learned part-level templates. Template colors indicate identity. Part bounding box colors indicate the chosen template.}
    \label{fig:2d_bbox}
\end{figure}

\subsection{Scene decomposition}

\begin{figure}[t]
    \centering \includegraphics[width=0.95\columnwidth]{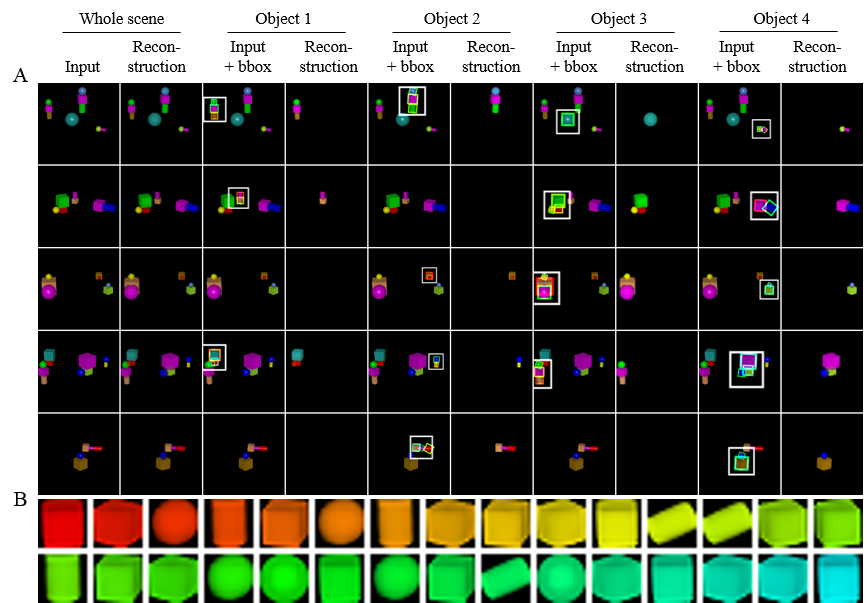}
    \caption{Qualitative results on 3D dataset. {\em (A)} Each row shows the overall reconstruction, and the predicted bounding boxes and reconstruction from each object cell for a given input image. {\em (B)} Learned part-level templates. Template colors indicate identity. Part bounding box colors indicate the chosen template.}
    \label{fig:3d_bbox}
\end{figure}

RICH is able to give interpretable, tree-structured decomposition of scenes into objects and parts. We visualize such decomposition in Figure \ref{fig:2d_bbox} and Figure \ref{fig:3d_bbox} for 2D and 3D scenes respectively, where we also show the learned memory templates for parts. Notice that the templates should be in gray scale, but for visualization purposes we have assigned a color to each template. The bounding boxes are drawn on top of the input images, according to the inferred pose of objects and parts with $\rz^\pres=1$. Object bounding boxes are drawn in white, while part bounding boxes are drawn in color to indicate the template chosen for each part.

We find that the templates have learned the appearance of parts at several canonical poses (rotation in 2D and viewpoint in 3D), and RICH predicts the pose of parts with respect to these canonical poses. This makes the decomposition even more interpretable. Moreover, equipped with templates, RICH is able to correctly identify the parts even when they have severe occlusion. See Figure \ref{fig:3d_bbox}A third row where a ball occludes an equally sized cube. This example (and many others) also demonstrate that RICH can successfully deal with objects composed of multiple parts that are of the same type.

\begin{table}[t]
	\centering
	\caption{Quantitative results on object and part detection.}
	\label{tbl:acc}
  	\begin{tabular}{ccccccc}
      \toprule
      Dataset & \multicolumn{3}{c}{2D dataset} & \multicolumn{3}{c}{3D dataset} \\
      \cmidrule(r){1-1} \cmidrule(r){2-4}  \cmidrule(r){5-7}
      
      Training set & 1$\sim$4 objects & \multicolumn{2}{c}{1\&3 objects} & 1$\sim$4 objects & \multicolumn{2}{c}{1\&3 objects} \\
      \cmidrule(r){1-1} \cmidrule(r){2-2} \cmidrule(r){3-4} \cmidrule(r){5-5} \cmidrule(r){6-7}
      
      Test set & 1$\sim$4 objects & 2 objects & 4 objects & 1$\sim$4 objects & 2 objects & 4 objects \\
      \cmidrule(r){1-1} \cmidrule(r){2-2} \cmidrule(r){3-3} \cmidrule(r){4-4} \cmidrule(r){5-5} \cmidrule(r){6-6} \cmidrule(r){7-7}
      \midrule
  	  \multicolumn{1}{l}{Object count error} & 0.00083 & 0.0014 & 0.00036 & 0.094 & 0.26 & 0.47 \\
      \multicolumn{1}{l}{Object precision} & 0.9985 & 0.9987 & 0.9996 & 0.9639 & 0.9157 & 0.9581 \\
      \multicolumn{1}{l}{Object recall} & 0.9984 & 0.9982 & 0.9995 & 0.9597 & 0.9758 & 0.8462 \\ \midrule
      \multicolumn{1}{l}{Part count error} & 0.0086 & 0.011 & 0.014 & 0.80 & 1.1 & 1.3 \\
      \multicolumn{1}{l}{Part precision} & 0.9991 & 0.9988 & 0.9991 & 0.8282 & 0.7579 & 0.8100 \\
      \multicolumn{1}{l}{Part recall} & 0.9989 & 0.9985 & 0.9993 & 0.9116 & 0.9258 & 0.8347 \\ \bottomrule
  	\end{tabular}
\end{table}

In addition to part-level templates, we believe that the learned object-level $\rvz_{O}^\appr$ also helps scene decomposition, especially when there is ambiguity in part assignment and occlusion between objects. For example, in Figure \ref{fig:2d_bbox}A first column, the triangle and circle near the center are close to each other and may well constitute an object. However, because this pose configuration is relatively rare in the training set (compare second column), RICH has correctly rejected this composition and instead assigned these two parts to separate objects, which better agrees with the training distribution. In Figure \ref{fig:3d_bbox}A fourth row, object 1 is occluded by object 3. RICH has successfully detected object 1 and added in the reconstruction a ball of the same color as the occluded part. This is quite reasonable since the augmented object is one of our predefined types and appears frequently in the dataset.

To quantify RICH's ability of scene decomposition and representation learning, we report absolute counting error, precision, and recall for detection of objects and parts in Table \ref{tbl:acc}, and compare the negative log-likelihood of RICH with a VAE \citep{kingma2013auto} baseline in Table \ref{tbl:nll}. Here the counting error measures the absolute difference between the predicted and true number of objects and parts. To obtain precision and recall, we need to match the predictions with the groundtruth. We set the matching priority as the distance between the predicted and true center positions, namely closer pairs of prediction and groundtruth will be matched first. We only match the pair if their distance is less than 10 pixels (less than half of the size of large parts). This ensures that the matched predictions will have approximately correct center positions. The VAE baseline shares the same scene-level encoder with RICH, and uses sub-pixel convolution \citep{shi2016real} for the decoder. We approximate the negative log-likelihood using 50 importance-weighted samples. The counting error, precision, and recall are also averaged over 50 samples from the posterior. As can be seen from Table \ref{tbl:acc}, RICH gives almost perfect detection of objects and parts on 2D dataset, and still performs reasonably well on the challenging 3D dataset. We observe that RICH tends to split a long cylinder into two parts, leading to the drop in precision for parts.

\begin{figure}[t]
    \centering \includegraphics[width=0.95\columnwidth]{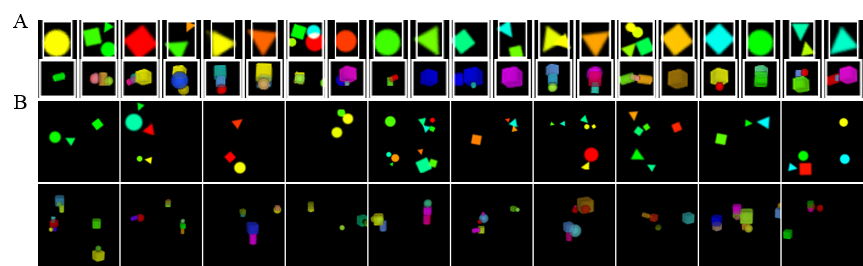}
    \caption{{\em (A)} Generated objects on {\em (Top)} 2D dataset and {\em (Bottom)} 3D dataset. White boxes indicate aspect ratio, and are drawn according to $\rvz_{O}^\ratio$. {\em (B)} Generated scenes on {\em (Top)} 2D dataset and {\em (Bottom)} 3D dataset.}
    \label{fig:gen}
\end{figure}

\begin{table}[t]
	\centering
	\caption{Comparison on negative log-likelihood.}
	\label{tbl:nll}
  	\begin{tabular}{ccccccc}
      \toprule
      Dataset & \multicolumn{3}{c}{2D dataset} & \multicolumn{3}{c}{3D dataset} \\
      \cmidrule(r){1-1} \cmidrule(r){2-4}  \cmidrule(r){5-7}

      Training set & 1$\sim$4 objects & \multicolumn{2}{c}{1\&3 objects} & 1$\sim$4 objects & \multicolumn{2}{c}{1\&3 objects} \\
      \cmidrule(r){1-1} \cmidrule(r){2-2} \cmidrule(r){3-4} \cmidrule(r){5-5} \cmidrule(r){6-7}

      Test set  & 1$\sim$4 objects   & 2 objects   & 4 objects   & 1$\sim$4 objects  & 2 objects  & 4 objects   \\
      \cmidrule(r){1-1} \cmidrule(r){2-2} \cmidrule(r){3-3} \cmidrule(r){4-4} \cmidrule(r){5-5} \cmidrule(r){6-6} \cmidrule(r){7-7}
      \midrule
  	  \multicolumn{1}{c}{VAE} & -13761.9 & -13801.4 & -13590.5 & -13712.0 & -13788.3 & -13433.2 \\
      \multicolumn{1}{c}{RICH} & \textbf{-13890.3}   & \textbf{-13908.3}   & \textbf{-13796.6}   & \textbf{-13818.0}   & \textbf{-13867.3}  & \textbf{-13539.7}   \\ \bottomrule
  	\end{tabular}
\end{table}

\subsection{Object and scene generation}

Apart from learning the part-level templates, RICH also has the ability to generate objects and scenes by recursively composing the learned templates. We show generation results in Figure \ref{fig:gen}. To generate the scenes, we first sample $\rvz_{S}^\appr \sim \mathcal{N}(\bm{0}, \rmI)$, and then sample other latents following the learned conditional prior distributions, and finally use the decoder to render the image. The objects are generated similarly, except that we decode up to the object level and ignore $\rvz_{O}^\pres$ and $\rvz_{O}^\pose$. We find that RICH has captured many predefined object types in the dataset, and also managed to come up with novel compositions. The generated scenes are also reasonable, with moderate distance and occlusion between objects.

\subsection{Generalization performance}

\begin{figure}[t]
    \centering \includegraphics[width=0.95\columnwidth]{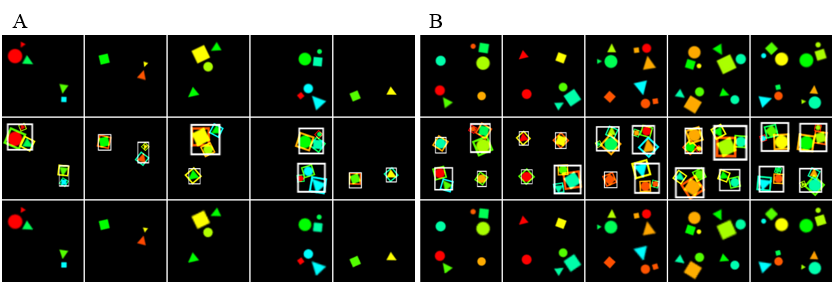}
    \caption{Generalization results on 2D dataset. RICH has been trained on scenes with 1 and 3 objects only, and tested on scenes with {\em (A)} 2 objects and {\em (B)} 4 objects. {\em (Top)} Input image. {\em (Middle)} Input image superimposed with predicted bounding boxes. {\em (Bottom)} Reconstruction.}
    \label{fig:2d_gen}
\end{figure}

\begin{figure}[t]
    \centering \includegraphics[width=0.95\columnwidth]{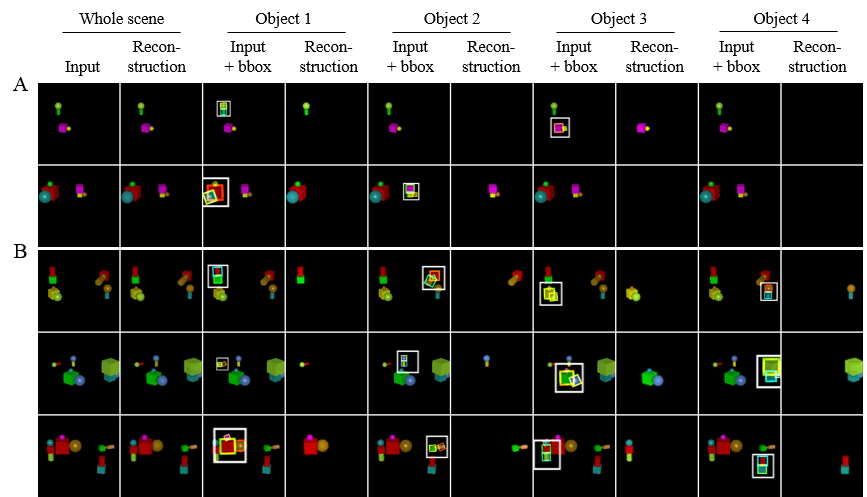}
    \caption{Generalization results on 3D dataset. RICH has been trained on scenes with 1 and 3 objects only, and tested on scenes with {\em (A)} 2 objects and {\em (B)} 4 objects. Each row shows the overall reconstruction, and the predicted bounding boxes and reconstruction from each object cell for a given input image.}
    \label{fig:3d_gen}
\end{figure}

RICH represents the scene as composition of objects and parts. This naturally enables generalization to novel scenes. Here we evaluate RICH's capacity to generalize to scenes with novel number of objects. The training and validation sets of this task contain scenes of one and three objects only. We trained RICH and the VAE baseline again, and report the metrics in Table \ref{tbl:acc} and Table \ref{tbl:nll} on two test sets, one having two-object scenes only, and the other having four-object scenes only. We also show qualitative results in Figure \ref{fig:2d_gen} and Figure \ref{fig:3d_gen}. As can be seen, RICH demonstrates quite decent generalization performance in both 2D and 3D scenes. We notice that in 3D case, there is a drop in recall when RICH is tested on four-object scenes. One reason is that four-object scenes exhibit more severe occlusion than the training set, and we have observed that when two objects are close and have occlusion, RICH would sometimes merge them into one object. Another reason is that in four-object scenes, objects are more likely to be partially outside the scene. In this case, RICH has difficulty predicting the precise object position, leading to unmatched predictions when we compute the recall.

\section{Conclusion}

We have proposed RICH, the first hierarchical generative model for learning interpretable compositional structures. RICH takes a top-down approach to infer the probabilistic scene graph representation for a given scene. This utilizes the higher-level appearance information to guide lower-level decomposition, thus avoiding the difficult routing problem faced by bottom-up approaches. Through extensive experiments, we have demonstrated that RICH is able to learn the compositional hierarchy from images containing multiple objects. An interesting future direction is to extend RICH to the sequential setting.

\subsubsection*{Acknowledgments}
SA thanks to Kakao Brain, Center for Super Intelligence (CSI), and Element AI for their support.

\bibliography{iclr2020_conference}

\begin{thebibliography}{24}
\providecommand{\natexlab}[1]{#1}
\providecommand{\url}[1]{\texttt{#1}}
\expandafter\ifx\csname urlstyle\endcsname\relax
  \providecommand{\doi}[1]{doi: #1}\else
  \providecommand{\doi}{doi: \begingroup \urlstyle{rm}\Url}\fi

\bibitem[Bachman(2016)]{bachman2016architecture}
Philip Bachman.
\newblock An architecture for deep, hierarchical generative models.
\newblock In \emph{Advances in Neural Information Processing Systems}, pp.\
  4826--4834, 2016.

\bibitem[Bornschein et~al.(2017)Bornschein, Mnih, Zoran, and
  Rezende]{bornschein2017variational}
J{\"o}rg Bornschein, Andriy Mnih, Daniel Zoran, and Danilo~Jimenez Rezende.
\newblock Variational memory addressing in generative models.
\newblock In \emph{Advances in Neural Information Processing Systems}, pp.\
  3920--3929, 2017.

\bibitem[Crawford \& Pineau(2019)Crawford and Pineau]{crawford2019spatially}
Eric Crawford and Joelle Pineau.
\newblock Spatially invariant unsupervised object detection with convolutional
  neural networks.
\newblock In \emph{Proceedings of AAAI}, 2019.

\bibitem[Deng et~al.(2019)Deng, Kornblith, and Hinton]{deng2019cerberus}
Boyang Deng, Simon Kornblith, and Geoffrey Hinton.
\newblock Cerberus: A multi-headed derenderer.
\newblock \emph{arXiv preprint arXiv:1905.11940}, 2019.

\bibitem[Eslami et~al.(2016)Eslami, Heess, Weber, Tassa, Szepesvari, Hinton,
  et~al.]{eslami2016attend}
SM~Ali Eslami, Nicolas Heess, Theophane Weber, Yuval Tassa, David Szepesvari,
  Geoffrey~E Hinton, et~al.
\newblock Attend, infer, repeat: Fast scene understanding with generative
  models.
\newblock In \emph{Advances in Neural Information Processing Systems}, pp.\
  3225--3233, 2016.

\bibitem[Foley et~al.(1996)Foley, Van, Van~Dam, Feiner, Hughes, Hughes, and
  Angel]{foley1996computer}
James~D Foley, Foley~Dan Van, Andries Van~Dam, Steven~K Feiner, John~F Hughes,
  J~Hughes, and Edward Angel.
\newblock \emph{Computer graphics: principles and practice}, volume 12110.
\newblock Addison-Wesley Professional, 1996.

\bibitem[Hinton et~al.(2018)Hinton, Sabour, and Frosst]{e2018matrix}
Geoffrey~E Hinton, Sara Sabour, and Nicholas Frosst.
\newblock Matrix capsules with {EM} routing.
\newblock In \emph{International Conference on Learning Representations}, 2018.
\newblock URL \url{https://openreview.net/forum?id=HJWLfGWRb}.

\bibitem[Jaderberg et~al.(2015)Jaderberg, Simonyan, Zisserman,
  et~al.]{jaderberg2015spatial}
Max Jaderberg, Karen Simonyan, Andrew Zisserman, et~al.
\newblock Spatial transformer networks.
\newblock In \emph{Advances in neural information processing systems}, pp.\
  2017--2025, 2015.

\bibitem[Jang et~al.(2016)Jang, Gu, and Poole]{jang2016categorical}
Eric Jang, Shixiang Gu, and Ben Poole.
\newblock Categorical reparameterization with gumbel-softmax.
\newblock \emph{arXiv preprint arXiv:1611.01144}, 2016.

\bibitem[Kingma \& Welling(2013)Kingma and Welling]{kingma2013auto}
Diederik~P Kingma and Max Welling.
\newblock Auto-encoding variational bayes.
\newblock \emph{arXiv preprint arXiv:1312.6114}, 2013.

\bibitem[Lee et~al.(2009)Lee, Grosse, Ranganath, and Ng]{lee2009convolutional}
Honglak Lee, Roger Grosse, Rajesh Ranganath, and Andrew~Y Ng.
\newblock Convolutional deep belief networks for scalable unsupervised learning
  of hierarchical representations.
\newblock In \emph{Proceedings of the 26th annual international conference on
  machine learning}, pp.\  609--616. ACM, 2009.

\bibitem[Maal{\o}e et~al.(2019)Maal{\o}e, Fraccaro, Li{\'e}vin, and
  Winther]{maaloe2019biva}
Lars Maal{\o}e, Marco Fraccaro, Valentin Li{\'e}vin, and Ole Winther.
\newblock Biva: A very deep hierarchy of latent variables for generative
  modeling.
\newblock \emph{arXiv preprint arXiv:1902.02102}, 2019.

\bibitem[Maddison et~al.(2016)Maddison, Mnih, and Teh]{maddison2016concrete}
Chris~J Maddison, Andriy Mnih, and Yee~Whye Teh.
\newblock The concrete distribution: A continuous relaxation of discrete random
  variables.
\newblock \emph{arXiv preprint arXiv:1611.00712}, 2016.

\bibitem[Romaszko et~al.(2017)Romaszko, Williams, Moreno, and
  Kohli]{romaszko2017vision}
Lukasz Romaszko, Christopher~KI Williams, Pol Moreno, and Pushmeet Kohli.
\newblock Vision-as-inverse-graphics: Obtaining a rich 3d explanation of a
  scene from a single image.
\newblock In \emph{Proceedings of the IEEE International Conference on Computer
  Vision}, pp.\  851--859, 2017.

\bibitem[Sabour et~al.(2017)Sabour, Frosst, and Hinton]{sabour2017dynamic}
Sara Sabour, Nicholas Frosst, and Geoffrey~E Hinton.
\newblock Dynamic routing between capsules.
\newblock In \emph{Advances in neural information processing systems}, pp.\
  3856--3866, 2017.

\bibitem[Shi et~al.(2016)Shi, Caballero, Husz{\'a}r, Totz, Aitken, Bishop,
  Rueckert, and Wang]{shi2016real}
Wenzhe Shi, Jose Caballero, Ferenc Husz{\'a}r, Johannes Totz, Andrew~P Aitken,
  Rob Bishop, Daniel Rueckert, and Zehan Wang.
\newblock Real-time single image and video super-resolution using an efficient
  sub-pixel convolutional neural network.
\newblock In \emph{Proceedings of the IEEE conference on computer vision and
  pattern recognition}, pp.\  1874--1883, 2016.

\bibitem[S{\o}nderby et~al.(2016)S{\o}nderby, Raiko, Maal{\o}e, S{\o}nderby,
  and Winther]{sonderby2016ladder}
Casper~Kaae S{\o}nderby, Tapani Raiko, Lars Maal{\o}e, S{\o}ren~Kaae
  S{\o}nderby, and Ole Winther.
\newblock Ladder variational autoencoders.
\newblock In \emph{Advances in neural information processing systems}, pp.\
  3738--3746, 2016.

\bibitem[Tieleman(2014)]{tieleman2014optimizing}
Tijmen Tieleman.
\newblock \emph{Optimizing neural networks that generate images}.
\newblock University of Toronto (Canada), 2014.

\bibitem[Todorov et~al.(2012)Todorov, Erez, and Tassa]{todorov2012mujoco}
Emanuel Todorov, Tom Erez, and Yuval Tassa.
\newblock Mujoco: A physics engine for model-based control.
\newblock In \emph{2012 IEEE/RSJ International Conference on Intelligent Robots
  and Systems}, pp.\  5026--5033. IEEE, 2012.

\bibitem[Watters et~al.(2019)Watters, Matthey, Burgess, and
  Lerchner]{watters2019spatial}
Nicholas Watters, Loic Matthey, Christopher~P Burgess, and Alexander Lerchner.
\newblock Spatial broadcast decoder: A simple architecture for learning
  disentangled representations in vaes.
\newblock \emph{arXiv preprint arXiv:1901.07017}, 2019.

\bibitem[Wu et~al.(2017)Wu, Tenenbaum, and Kohli]{wu2017neural}
Jiajun Wu, Joshua~B Tenenbaum, and Pushmeet Kohli.
\newblock Neural scene de-rendering.
\newblock In \emph{Proceedings of the IEEE Conference on Computer Vision and
  Pattern Recognition}, pp.\  699--707, 2017.

\bibitem[Yao et~al.(2018)Yao, Hsu, Zhu, Wu, Torralba, Freeman, and
  Tenenbaum]{yao20183d}
Shunyu Yao, Tzu~Ming Hsu, Jun-Yan Zhu, Jiajun Wu, Antonio Torralba, Bill
  Freeman, and Josh Tenenbaum.
\newblock 3d-aware scene manipulation via inverse graphics.
\newblock In \emph{Advances in Neural Information Processing Systems}, pp.\
  1887--1898, 2018.

\bibitem[Zhao et~al.(2017)Zhao, Song, and Ermon]{zhao2017learning}
Shengjia Zhao, Jiaming Song, and Stefano Ermon.
\newblock Learning hierarchical features from deep generative models.
\newblock In \emph{Proceedings of the 34th International Conference on Machine
  Learning-Volume 70}, pp.\  4091--4099. JMLR. org, 2017.

\bibitem[Zhu et~al.(2008)Zhu, Lin, Huang, Chen, and
  Yuille]{zhu2008unsupervised}
Long~Leo Zhu, Chenxi Lin, Haoda Huang, Yuanhao Chen, and Alan Yuille.
\newblock Unsupervised structure learning: Hierarchical recursive composition,
  suspicious coincidence and competitive exclusion.
\newblock In \emph{European Conference on Computer Vision}, pp.\  759--773.
  Springer, 2008.

\end{thebibliography}
\bibliographystyle{iclr2020_conference}

\end{document}